%% file: main.tex
\documentclass[12pt]{article} 
\usepackage{amsmath,amsthm,amssymb,bm,mathrsfs}
\usepackage{graphicx,subfigure}
\usepackage{caption,enumerate,listings,setspace}
\usepackage[colorlinks,linkcolor=red,anchorcolor=blue,citecolor=blue]{hyperref}
\usepackage[margin=1in]{geometry}
\usepackage[title]{appendix}
\usepackage{enumitem}
\theoremstyle{plain}
\usepackage{booktabs}
\usepackage[numbers]{natbib}
\usepackage{array}
\usepackage{algorithm}
\usepackage{algpseudocode}
\usepackage{makecell}

\input{def-smile}

\input{front}

\title{\textbf{
\texttt{BadGD}: A unified data-centric framework to identify gradient descent vulnerabilities}} 

\author
{
Chi-Hua Wang \thanks{Postdoctoral Scholar, Department of Statistics and Data Science,  UCLA, CA, 90095. Email: chihuawang@ucla.edu},
Guang Cheng\thanks{Professor, Department of Statistics and Data Science, UCLA, CA, 90095. Email: guangcheng@ucla.edu}
}

\begin{document} 

\maketitle

\begin{abstract}
We present \texttt{BadGD}, a unified theoretical framework that exposes the vulnerabilities of gradient descent algorithms through strategic backdoor attacks. Backdoor attacks involve embedding malicious triggers into a training dataset to disrupt the model’s learning process. Our framework introduces three novel constructs: Max RiskWarp Trigger, Max GradWarp Trigger, and Max GradDistWarp Trigger, each designed to exploit specific aspects of gradient descent by distorting empirical risk, deterministic gradients, and stochastic gradients respectively. We rigorously define clean and backdoored datasets and provide mathematical formulations for assessing the distortions caused by these malicious backdoor triggers. By measuring the impact of these triggers on model training procedure, our framework bridges existing empirical findings with theoretical insights, demonstrating how malicious party can exploit gradient descent hyperparameters to maximize attack effectiveness. In particular, we show that these exploitation can significantly alter the loss landscape and gradient calculations, leading to compromised model integrity and performance. This research underscores the severe threats posed by such data-centric attacks and highlights the urgent need for robust defenses in machine learning. \texttt{BadGD} sets a new standard for understanding and mitigating adversarial manipulations, ensuring the reliability and security of AI systems.
\end{abstract}


\clearpage
\section{Introduction}
\label{sec:intro}

Gradient Descent is vulnerable to adversarial examples. It has been observed that a single crafted adversarial example can make substantial gradient distortion \cite{jagielski2020auditing}, which makes \textit{backdoor attacks} \cite{gu2017badnets, saha2020hidden} become main principal auditing tool to test and audit differential privacy guarantee of machine learning training procedure. Utilizing this vulnerability, researchers have developed some white-box privacy attack methods \cite{nasr2023tight}, in which they first obtain the training routine of a victim model and then use the information to craft the adversarial example (backdoor trigger) to make most effective distortion of training process to invalidate the differential privacy guarantee. Canary attacks is the name that the literature of differential private machine learning called for the backdoor attack.

Backdoor attacks exhibit many intriguing properties. For instance, Nasr et al. \cite{nasr2023tight} found that the adversarial examples crafted using partial knowledge of the training routine can  achieve a significant unhappy impact on the training process. 
Previous works have attempted to explain these properties from different perspectives \cite{gu2017badnets, saha2020hidden, jagielski2020auditing}. Unfortunately, these works only provide \textbf{ad hoc} explanations for the properties of backdoor attacks without quantitative analyses under a unified framework. For example, Jagielski et al. \cite{jagielski2020auditing} suggest that the effectiveness of backdoor attacks is due to the adversarial examples exploiting specific vulnerabilities about gradient direction in the training routine, but the properties of these vulnerabilities are not discussed quantitatively. As for the cause of the gradient distortion, a plausible explanation is that the adversarial examples introduce non-random, targeted perturbations that amplify the gradient errors during training \cite{jagielski2020auditing}. However, these targeted perturbations are hard to characterize, and it is even more challenging to associate them with the properties of backdoor attacks. More importantly, this explanation contradicts some experimental results by Nasr et al. \cite{nasr2023tight}, which show that adversarial examples can still significantly distort gradients even when the perturbations are not perfectly targeted. We refer the readers to Section \ref{sec:related_works} for more related works. In summary, existing analyses make good sense within their respective settings, but it is unclear whether they are in accord with each other. It is natural to ask the following question: 

\begin{center}
\textit{Can we form a unified theoretical framework that explains the properties of backdoor attacks?}
\end{center}

\begin{figure}[t]
    \centering
    \includegraphics[width = 1.00\linewidth]{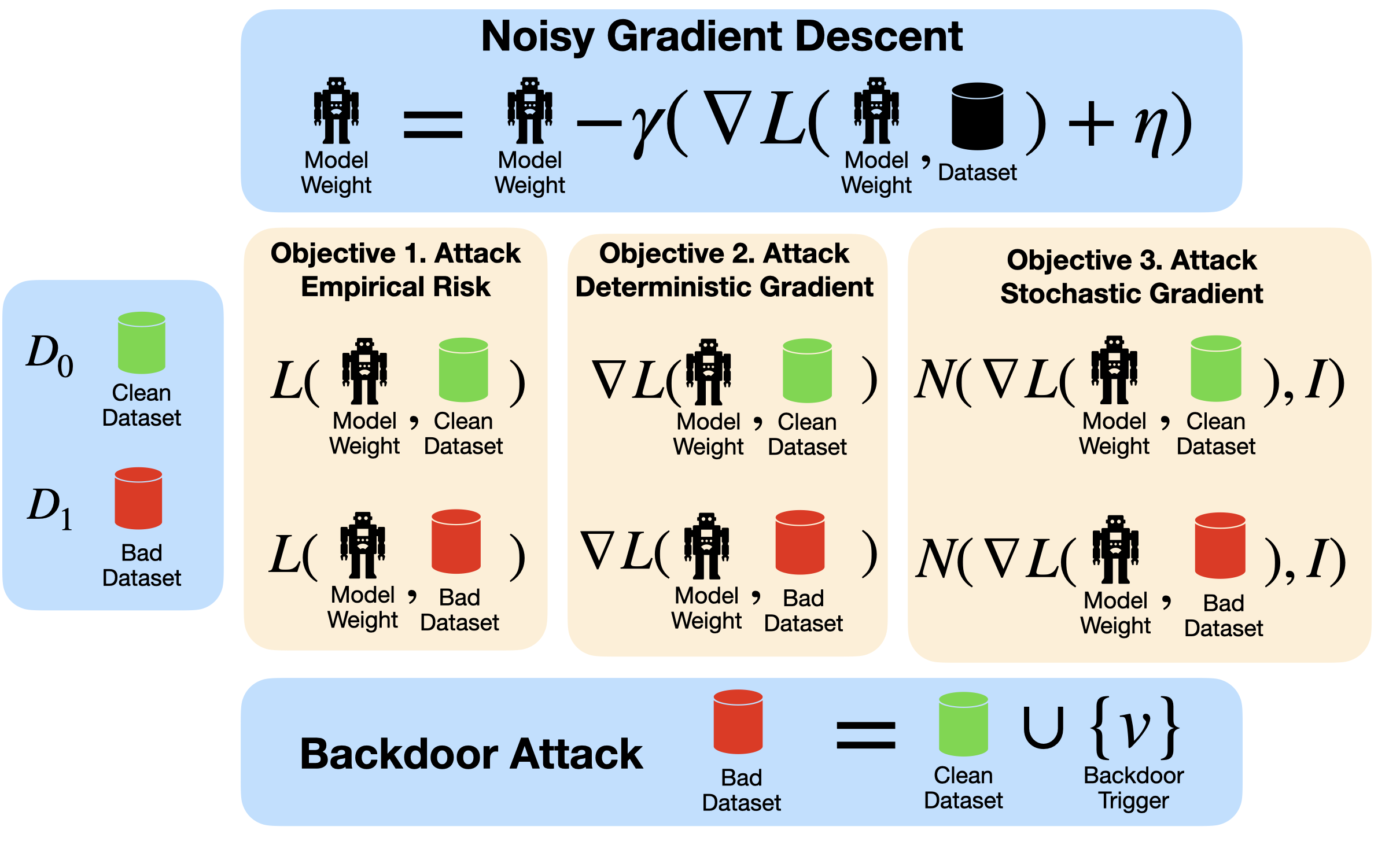}
    \caption{\texttt{BadGD} attack model: the malicious user crafts a backdoor trigger and adds into clean dataset to construct a bad dataset. The aim of the malicious user is to maximize the distortion of attack objectives. Section \ref{sec:BadGD_attack_model} devotes to use bad dataset to distort empirical risk and both deterministic and stochastic gradients. 
    Section \ref{sec:attack_supervised_learning} applies the \texttt{BadGD} attack model to identify the vulnerabilities of gradient descent when training  supervised learning models with square loss. 
    }
\label{fig:badgd_attack_model}

\end{figure}

This paper gives a positive answer to this question. More specifically, we propose an attack model, called the \textit{BadGD} framework (Figure \ref{fig:badgd_attack_model}), that formalizes popular beliefs and explains the existing empirical results in a unified theoretical framework. Our model assumes that the training data can be clean or backdoored and that backdoor triggers can be strategically designed to exploit specific vulnerabilities in gradient descent algorithms. The central part of our model introduces three novel constructs: Max RiskWarp Trigger (Def \ref{def:max_riskwarp_trigger}), Max GradWarp Trigger (Def \ref{def:max_gradwarp_trigger}), and Max GradDistWarp Trigger (Def \ref{def:max_GradDistWarp_trigger}). These constructs are motivated by the need to assess and quantify the distortions caused by malicious backdoor triggers on empirical risk, deterministic gradients, and stochastic gradients, respectively. Such constructs matches our goal  to provide a systematic approach to understand and mitigate the threats posed by such adversarial manipulations.

In summary, we propose an attack model that is theoretically tractable and consistent with the focus of existing results. Our attack models formalizes widespread beliefs and explains the properties of gradient descent vulnerabilities under a unified framework. The remainder of this paper is structured as follows. Sections \ref{sec:related_works} and \ref{sec:problem_setup} introduce the related works and basic problem formulation, respectively. Section \ref{sec:BadGD_attack_model} presents our \texttt{BadGD} attack models and Section \ref{sec:attack_supervised_learning} applies our attack model in supervised learning setting to identify the vulnerabilities of Gradient Descent. Section \ref{sec:conclusion} summarizes this paper.

\section{Relate Works}
\label{sec:related_works}

\textbf{Noisy gradient descent.} Our research contributes to the machine learning security community, with a particular focus on Noisy Gradient Descent (Noisy GD) \cite{das2023beyond, avella2023differentially, tholeti2021differentially}. Noisy Gradient Descent (Noisy GD) is best exemplified by DP-SGD \cite{abadi2016deep}, which introduces noise into gradient updates to ensure privacy protection. In the DP-SGD framework, each update incurs a privacy budget, which quantifies the privacy cost of the operation. However, this approach is not without vulnerabilities. In Sec \ref{sec:BadGD_attack_model}, we introduce a novel backdoor trigger that exploits these vulnerabilities. Our backdoor trigger strategically amplifies the privacy budget required for each update (Lemma \ref{lm:privacy_budget_graddistwarp_linear_regression}), leading to substantial disruption in the training process. This mechanism effectively increases the noise level in the gradient, distorting the learning trajectory and undermining the model's integrity. Our findings highlight the critical need for more robust defenses against such sophisticated attacks in privacy-preserving machine learning paradigms.

\textbf{Privacy auditing.}
Our \texttt{BadGD} attack model (Fig. \ref{fig:badgd_attack_model}) is inspired by seminal research on Privacy Auditing \cite{jagielski2020auditing}, a process that empirically tests the theoretical privacy guarantees in differentially private machine learning \cite{bassily2014private, wang2017differentially, zhang2017efficient, bassily2019private, koloskova2023revisiting}. Privacy Auditing seeks to verify whether the privacy budget upper bound, which is promised by theory, holds true in practice \cite{koskela2020computing, nasr2023tight, Kong2023DPAuditoriumAL, pillutla2024unleashing, kazmi2024panoramia}. A pivotal element of this verification involves selecting a poisoning point—termed as a \textit{backdoor trigger} in our work—to execute the auditing task. However, previous methods for selecting these poisoning points have been largely \textbf{ad hoc}, leading to inconsistent and unreliable results. Our research addresses this critical gap by providing a systematic approach to constructing backdoor triggers (Sec \ref{sec:distor_stoc_gradient_bad_dataset}). This innovative method enhances the reliability of Privacy Auditing, ensuring that the privacy budget is tested rigorously and consistently. By offering a structured and replicable technique for creating these triggers, our \texttt{BadGD} framework not only strengthens the empirical validation of differential privacy guarantees but also sets a new standard for future research in this domain. We believe our contribution will significantly improve the theoretical tools available for Privacy Auditing, fostering greater trust and robustness in privacy-preserving machine learning models.

\textbf{Gaussian differential privacy.} The framework of Gaussian Differential Privacy (GDP) \cite{dong2021gaussian, balle2018improving} serves as a critical tool for quantifying the privacy risks posed by backdoor attacks. When attacking Stochastic Gradient Descent (SGD) \cite{garrigos2023handbook}, the updates made to the machine learning model are essentially random variables. The GDP framework allows us to measure the differences in these updates before and after an attack (Lemma \ref{lm:stoc_gd_gap_clean_bad}), thus providing a clear picture of the attack's impact. Specifically, GDP helps us calculate the divergence introduced by our backdoor attack, quantifying the additional privacy budget expenditure required (Lemma \ref{lm:privacy_budget_graddistwarp_linear_regression}). This measurement is crucial as it enables us to assess the extent of vulnerability in Gradient Descent methods. By systematically increasing the privacy budget needed for each update, our backdoor trigger can cause significant disruptions, highlighting the need for robust defenses in privacy-preserving machine learning paradigms.

Here we recite two concepts in \cite{dong2021gaussian} to prepare our discussion on the privacy concern rooted from  backdoor attacks. The first concept is tradeoff function: for any two probability distributions $P$ and $Q$ on the same space, define the trade-off function $T(P, Q):[0,1] \rightarrow[0,1]$ as
\begin{equation}\label{eq:tradeoff_function}
T(P, Q)(\alpha)=\inf \left\{\beta_\phi: \alpha_\phi \leqslant \alpha\right\},
\end{equation}
where the infimum is taken over all (measurable) rejection rules. The second concept is Gaussian Differential Private (GDP) mechanism: a mechanism is \textit{$\mu-GDP$} if and only if it is $(\varepsilon, \delta(\varepsilon))-DP$ \cite{dwork2006calibrating, wasserman2010statistical} for all $\varepsilon \geqslant 0$, where
\begin{equation}\label{eq:GDP_iff_DP}
\delta(\varepsilon)=\Phi\left(-\frac{\varepsilon}{\mu}+\frac{\mu}{2}\right)-\mathrm{e}^{\varepsilon} \Phi\left(-\frac{\varepsilon}{\mu}-\frac{\mu}{2}\right).
\end{equation}

Here, $\epsilon$ is called \textit{privacy budget}. Intuitively, if we can provide a backdoor trigger such that a privacy budget for single update is huge, then such trigger is an effective backdoor attack since it raises serious concerns about training data confidentiality.

\section{Problem Setup}
\label{sec:problem_setup}

In this section, we prepare reader essential notations and formulations to appreciate the proposed \texttt{BadGD} attack model established at Section \ref{sec:BadGD_attack_model}. Section \ref{subsec:Emp_GD} reviews the notion of empirical risk and gradient descent. Section \ref{subsec:attack_objectives} describes the attack objectives that allows malicious users to distort machine learning training process. Section \ref{subsec:attack_data_poisoning} formally defines the clean and bad dataset, and their empirical risk and gradients.

\subsection{Notations}
\label{subsec:Emp_GD}

Minimizing the empirical risk is essential in training machine learning models—it directly enhances the model’s capacity to fit the training data with precision, acting as the primary compass guiding the optimization of model parameters. Formally, we defines the empirical as follows:
\begin{definition}[Empirical Risk] \label{def:emp_risk}
Given a model weight $w$ and a dataset $D = \{(x_i, y_i)\}_{i=1}^{n}$, where $x_i \in \mathbb{R}^{d}$ is the feature vector and $y_{i} \in \mathbb{R}$ is the response value. We define the \textit{empirical risk} of weight $w$ on dataset $D$ as 
\begin{equation}\label{eq:Empirical_Risk}
    L(w, D) = \frac{1}{n} \sum_{i=1}^n \ell(w ; (x_i, y_i)).
\end{equation}
Here, the loss $\ell(w ; (x_i, y_i))$ evaluates the discrepancy between predicted outcomes and actual targets.
\end{definition}

Reducing empirical risk improves model accuracy and emphasizes the need for robust validation to prevent overfitting, ensuring reliable performance on new data. However, a malicious user adding just one data point to the training set can significantly threaten the integrity of the training process, raising serious concerns about model vulnerability (See lemma \ref{lm:construct_max_riskwarp_trigger}).

To minimize empirical risk \eqref{eq:Empirical_Risk}, we use \textit{Gradient Descent} (\texttt{GD}) 
and \textit{Noisy Gradient Descent} (\texttt{Noisy-GD})
. The following defines these key optimization models for updating the model weight \(w\):

\begin{definition}[Gradient Descent]
\label{def:GD}
Let $w \in \mathbb{R}^{d}$ be the model weight at latest checkpoint, $\gamma >0$ to be the learning rate and $D$ to be the dataset used to update the model. Let $L(w; D)$ denotes the empirical risk of model weight calculated on dataset $D$ as Definition \ref{def:emp_risk}. Then,  the \textit{Gradient Descent} (\texttt{GD}) algorithm is a training algorithm that update the model weight $w$ by the following iteration scheme:
\begin{equation}\label{eq:GD}
    w = w - \gamma \nabla L(w, D).
\end{equation}
\end{definition}

\begin{definition}[Noisy Gradient Descent] \label{def:noisy_GD}
Let $w, \gamma, D, L(w; D)$ denote the model weight, learning rate, dataset, empirical risk as in Definition \ref{def:GD}. Let $\eta$ denote a random sample from the multivariate Gaussian distribution $N(0, \sigma^2I_{d})$, where $\sigma$ is the perturbation scale. Then,  the \textit{Noisy Gradient Descent} (\texttt{Noisy-GD}) algorithm is a training algorithm that update the model weight $w$ by the following iteration scheme:
\begin{equation}\label{eq:GD}
    w = w - \gamma (\nabla L(w, D) + \eta).
\end{equation}
\end{definition}

GD and Noisy-GD are essential optimization techniques for minimizing empirical risk, thereby enhancing model accuracy. GD iteratively updates the model weight $w$ using the gradient of the empirical risk, promoting consistent improvement. On the other hand, Noisy-GD incorporates random noise into the gradient updates, which helps in preventing overfitting and exploring a broader solution space . However, these methods also subject to the risk that a single malicious data point can distort the training process, raising significant concerns about the vulnerability and privacy leakage of machine learning models (see lemma \ref{def:max_gradwarp_trigger} and lemma \ref{lm:privacy_budget_graddistwarp_linear_regression}).

\subsection{Attack Objectives}
\label{subsec:attack_objectives}

Training procedure of machine learning models is vulnerable. Adversarial attacks in machine learning can target various components of a model's training process, each with potentially profound impacts. Specifically, our proposed \texttt{BadGD} attack models consider these attacks can be designed to manipulate:
\begin{enumerate}[leftmargin=*]
    \item \textbf{Empirical Risk $L(\omega, D)$}: 
    Attack Empirical Risk (Def \ref{def:emp_risk}). 
    This is the function mapping a dataset to a real number, representing the model's performance metric (loss) over the dataset. An attack targeting the empirical risk seeks to distort the loss landscape, potentially leading to suboptimal model training.
    
    \item \textbf{Deterministic Gradient $\nabla_{\omega}L(\omega, D)$}: 
    Attack Gradient Descent (Def \ref{def:GD}). 
    Here, the attack targets the function that maps the dataset to a \(d\)-dimensional constant vector, the gradient of the loss function. By manipulating this gradient, adversaries can direct the training process towards specific, less optimal regions of the parameter space.
    
    \item \textbf{Stochastic Gradient $\mathcal{N}(-\eta \nabla L(w, D), \eta^2 \Gamma^2)$}: 
    Attack Noisy Gradient Descent (Def \ref{def:noisy_GD}).
    Unlike the deterministic gradient, the stochastic gradient is targeted through a function mapping the dataset to a $d$-dimensional random vector. This introduces variability in the gradient estimates, which can be exploited by adversaries to inject noise and uncertainty into the training process, further degrading model performance.
\end{enumerate}

\subsection{Backdoor Attack as Data Poisoning Attack}
\label{subsec:attack_data_poisoning}

Assume we are working with two distinct datasets: $D_0$, referred to as the \textit{clean} dataset, and $D_1$, known as the \textit{backdoored} dataset. The table below (Table \ref{tab:adversarial_targets}) illustrates the profound effects that various types of adversarial targets can have on machine learning training processes when applied to these datasets, with a specific focus on manipulations of empirical risk and gradients. The adversary’s intent is not merely to alter the training process; their objective is to engineer the targets on $D_1$ such that there is a significant and detrimental deviation from those on $D_0$. This strategic manipulation poses a serious threat to the integrity and reliability of machine learning models, as it can drastically compromise their performance and decision-making processes.

\begin{table}[ht]
\centering
\begin{tabular}{|c|c|c|c|}
\hline
\textbf{Dataset} & \textbf{Empirical Risk} & \textbf{Deterministic Gradient} & \textbf{Stochastic Gradient} \\
\hline
\(D_0\) & \(L(w, D_0)\) & \(-\eta \nabla L(w, D_0)\) & \(\mathcal{N}(-\eta \nabla L(w, D_0), \eta^2 \Gamma^2)\) \\
\(D_1\) & \(L(w, D_1)\) & \(-\eta \nabla L(w, D_1)\) & \(\mathcal{N}(-\eta \nabla L(w, D_1), \eta^2 \Gamma^2)\) \\
\hline
\end{tabular}
\caption{Adversarial Targets and Their Effects on Machine Learning Models}
\label{tab:adversarial_targets}
\end{table}

As shown in Table~\ref{tab:adversarial_targets}, the effects of adversarial manipulation can vary significantly depending on whether the attack targets the empirical risk, the deterministic gradient, or the stochastic gradient. By analyzing these targeted attacks, we aim to develop more robust machine learning models that can withstand such adversarial interventions, ensuring reliability and accuracy in real-world applications.

Now, we define how attacker can construct a bad dataset from a clean dataset. 

\begin{definition}[Clean dataset and Bad dataset]
\label{def:clean_bad_dataset}
We use $D_0=\left\{\left(x_i, y_i\right)\right\}_{i=1}^n$ to denote a clean training dataset (clean dataset) and \begin{equation}
    D_1 \equiv D_0 \cup\{v\}
\end{equation}
denote a backdoored training dataset (bad dataset) with a \textit{backdoor trigger} $v=\left(x_v, y_v\right)$.
\end{definition}
Given Definition \ref{def:clean_bad_dataset}, we reinforce that the objective of this paper is to provide a systematic approach to construct the  backdoor trigger $v=\left(x_v, y_v\right)$ such that the empirical risk or the gradient descent will be distorted greatly. Our framework is called \texttt{BadGD}, and will be established at Section \ref{sec:BadGD_attack_model}.

\begin{definition}[Clean empirical risk and clean gradient]
\label{def:clean_emp_risk}
Given a clean dataset $D_0$ define in Definition \ref{def:clean_bad_dataset}. For a model weight $w$, let $\ell(w; (x_i, y_i))$ denote the individual loss of $w$ on the training example $(x_i, y_i)$. Then, the \textit{clean empirical risk} $L(w; D_0)$ is defined as
\begin{equation}\label{eq:clean_emp_risk}
    L\left(w ; D_0\right)=\frac{1}{n} \sum_{i=1}^n \ell\left(w ;\left(x_i, y_i\right)\right)
\end{equation}
In addition, $\nabla_{w} L(w; D_0)$ is called \textit{clean gradient} of model weight $w$ on dataset $D_0$. 
\end{definition}

\begin{definition}[Bad Empirical Risk and bad gradient]
\label{def:bad_emp_risk}
Given a bad dataset $D_1$ define in Definition \ref{def:clean_bad_dataset}. For a model weight $w$, let $\ell(w; (x_i, y_i))$ denote the individual loss of $w$ on the training example $(x_i, y_i)$. Then, the \textit{bad empirical risk} $L(w; D_1)$ is defined as
\begin{equation}\label{eq:bad_emp_risk}
   L\left(w ; D_0 \cup \{v\} \right)=\frac{1}{n+1}\left[\sum_{i=1}^n \ell\left(w ;\left(x_i, y_i\right)\right)+\ell\left(w ;\left(x_v, y_v\right)\right)\right]    
\end{equation}
$\nabla_{w} L(w; D_0 \cup {v})$ is called \textit{bad gradient} of model weight $w$ on dataset $D_0$ via backdoor trigger $v$. 
\end{definition}
Lemma \ref{lm:id_clean_bad_gradient} describes a basic identity between clean gradient  and bad gradient :
\begin{lemma}[Identity between clean and bad gradient]\label{lm:id_clean_bad_gradient}
Given a clean dataset $D_0=\left\{\left(x_i, y_i\right)\right\}_{i=1}^n$ and a backdoor trigger $v$. The relation between clean gradient and bad gradient is  
\begin{equation}\label{eq:id_clean_bad_gradient}
    \nabla_{w} L(w; D_0 \cup \{v\})
    = (1-\frac{1}{n+1})     \nabla_{w} L(w; D_0 ) + \frac{1}{n+1} \nabla_{w} \ell(w; x_v, y_v).
\end{equation}
\end{lemma}
\begin{proof}

The basic identity is a direct result from definition \ref{def:clean_emp_risk} and \ref{def:bad_emp_risk}.
\end{proof}

\section{The \texttt{BadGD} Attack Model}
\label{sec:BadGD_attack_model}

In this section, we formulate the attack models that the malicious users could adopt the design of bad dataset (Def \ref{def:clean_bad_dataset}) to distort machine learning training process. 
Section \ref{sec:distor_emp_risk_bad_dataset} propose the problem to find the Max
RiskWarp Trigger (Def \ref{def:max_riskwarp_trigger}) that maximizes the distortion of empirical risk via bad dataset. Analogously, Section \ref{sec:distor_det_gradient_bad_dataset} and \ref{sec:distor_stoc_gradient_bad_dataset} proposes problems to find the Max GradWarp Trigger (Def \ref{def:max_gradwarp_trigger}) and the Max GradDistWarp Trigger (Def \ref{def:max_GradDistWarp_trigger}) that maximizes the distortion of deterministic and noisy gradient via bad dataset.We further point out the link between the Max GradDistWarp Trigger and the Gaussian differential privacy, founding our \texttt{BadGD} attack models as an effective tool for the task of privacy auditing.

\subsection{Distort Empirical Risk via Bad Dataset}
\label{sec:distor_emp_risk_bad_dataset}

We define \textit{Max RiskWarp Trigger} as a backdoor trigger $v$ crafted to maximize empirical risk distortion, dramatically altering model risk assessments. Formally, 
\begin{definition}[Max RiskWarp Trigger]
\label{def:max_riskwarp_trigger}
A backdoor trigger $v = (x_v, y_v)$ is a Max RiskWarp Trigger $v^{*}(w)$ for the model weight $w$ if $v$ maximizes the difference between bad and clean empirical risk:
\begin{equation}\label{eq:max_riskwarp_trigger}
    v^*(w)=\arg \max _v\left\{L\left(w ; D_0 \cup\{v\}\right)-L\left(w ; D_0\right)\right\}
\end{equation}
\end{definition}

\begin{lemma} \label{lm:emp_risk_gap_clean_bad}
Denote the backdoor trigger $v = (x_v, y_v)$. Then the difference between clean  empirical
risk \eqref{eq:clean_emp_risk} and bad empirical risk \eqref{eq:bad_emp_risk}  satisfies
\begin{equation}\label{eq:emp_risk_gap_clean_bad}
L\left(w ; D_0 \cup\{v\}\right)-L\left(w ; D_0\right)=\frac{1}{n+1}\left[\ell\left(w ;\left(x_v, y_v\right)\right)-L\left(w ; D_0\right)\right]
\end{equation}
\end{lemma}

\begin{proof}

See section \ref{subsec:pf_emp_risk_gap_clean_bad}.
\end{proof}

Lemma \ref{lm:emp_risk_gap_clean_bad} reveals that the impact of adding a single backdoor trigger $v = (x_v, y_v)$ to the dataset $D_0$ is quantitatively measured by the difference in empirical risks with and without the trigger. Specifically, it is calculated as a function of the loss of the trigger point relative to the average loss over the clean dataset $D_0$. This equation provides a direct method to assess how a specific backdoor sample affects the model's performance.

To design an effective backdoor trigger $v = (x_v, y_v)$, we can derive from the equation \eqref{eq:emp_risk_gap_clean_bad} that maximizing the term $\ell\left(w ;\left(x_v, y_v\right)\right) - L\left(w ; D_0\right)$ will result in the greatest distortion in empirical risk. This approach suggests that to create a "Max RiskWarp Trigger" as defined, the trigger's data point $(x_v, y_v)$ should be crafted such that:
\begin{enumerate}[leftmargin=*]
\item \textbf{High Loss at Backdoor Trigger Point}: $\ell\left(w ;\left(x_v, y_v\right)\right)$ should be high. This can be achieved by choosing $(x_v, y_v)$ such that it lies in a region where the model $w$ performs poorly, or by setting $y_v$ to a value that is significantly mispredicted by $w$ when given $x_v$.

\item \textbf{Maximize the Gap}: The term $\ell\left(w ;\left(x_v, y_v\right)\right) - L\left(w ; D_0\right)$ should be maximized. This means that not only should the loss at the trigger point be high, but it should also significantly deviate from the average loss over the clean dataset.
\end{enumerate}

By focusing on these two criteria, we strategically design backdoor triggers that maximize the disruption in empirical risk, thereby providing insights into the vulnerabilities of the learning system to such data manipulations (See sec \ref{sec:construct_max_riskwarp_trigger}). This is crucial for developing more robust machine learning models that are resistant to adversarial attacks or manipulations.

\subsection{Distort Deterministic Gradient via Bad Dataset}
\label{sec:distor_det_gradient_bad_dataset}

We define \textit{Max GradWarp Trigger} as a backdoor trigger $v$ crafted to maximally distort gradient calculations, dramatically steering the model’s learning path. Formally, 
\begin{definition}[Max GradWarp Trigger]
\label{def:max_gradwarp_trigger}
A backdoor trigger $v = (x_v, y_v)$ is a Max GradWarp Trigger $v^{*}(w)$ for the model weight $w$ if $v$ maximizes the difference between bad and clean  gradient:
\begin{equation}\label{eq:max_gradwarp_trigger}
    v^*(w)=\arg \max_v \|\nabla_{w}L\left(w ; D_0 \cup\{v\}\right)-\nabla_{w}L\left(w ; D_0\right)\|_2
\end{equation}
\end{definition}

\begin{lemma} \label{lm:det_gd_gap_clean_bad}
Denote the backdoor trigger $v = (x_v, y_v)$. Then the difference between clean deterministic gradient
and bad deterministic gradient  satisfies
\begin{equation}\label{eq:det_gd_gap_clean_bad}
\nabla_{w}L\left(w ; D_0 \cup\{v\}\right)-\nabla_{w}L\left(w ; D_0\right)=\frac{1}{n+1}\left[\nabla_{w}\ell\left(w ;\left(x_v, y_v\right)\right)-\nabla_{w}L\left(w ; D_0\right)\right]
\end{equation}
\end{lemma}
\begin{proof}
This is a direct consequence of lemma \ref{lm:emp_risk_gap_clean_bad}. \end{proof}

Lemma \ref{lm:det_gd_gap_clean_bad} elucidates that the influence of introducing a backdoor trigger \( v = (x_v, y_v) \) into a dataset \( D_0 \) is quantifiable through the difference in the gradient calculations of the loss function, with and without the trigger. This difference directly informs how much the inclusion of the trigger shifts the gradient, potentially redirecting the model's learning trajectory significantly.

In crafting a "Max GradWarp Trigger" (Def \ref{def:max_gradwarp_trigger}), the objective is to choose a trigger point \( (x_v, y_v) \) such that the difference in the gradients, \(\nabla_{w}\ell\left(w ;\left(x_v, y_v\right)\right) - \nabla_{w}L\left(w ; D_0\right)\), is maximized. This maximization will lead to the most substantial deviation in the model's gradient path when the trigger is included in the training set. To achieve this, consider the following design principles:

\begin{enumerate}[leftmargin=*]
\item \textbf{High Gradient at Trigger Point}: Ensure that \(\nabla_{w}\ell\left(w ;\left(x_v, y_v\right)\right)\) is large. This could be achieved by selecting features \(x_v\) and labels \(y_v\) that are particularly challenging for the model to predict correctly, or by setting \(y_v\) to a label that is far from the model's current prediction for \(x_v\), thus inducing a large gradient.

\item \textbf{Maximize Gradient Deviation}: The difference \(\nabla_{w}\ell\left(w ;\left(x_v, y_v\right)\right) - \nabla_{w}L\left(w ; D_0\right)\) should be as large as possible. This can be engineered by not only having a high gradient at the trigger point but also by ensuring that this gradient starkly contrasts with the average gradient computed over the clean dataset \(D_0\).
\end{enumerate}

By focusing on these elements, we design backdoor triggers that most effectively distort the gradient calculations (Sec. \ref{sec:construct_max_gradwarp_trigger}), guiding the model’s learning in unintended, potentially harmful directions. This insight is critical for understanding how data can be weaponized against machine learning systems and underscores the need for mechanisms to detect and mitigate such attacks to ensure the integrity and robustness of machine learning models.

\subsection{Distort Stochastic Gradient via Bad Dataset}
\label{sec:distor_stoc_gradient_bad_dataset}

We define \textit{Max GradDistWarp Trigger} as a backdoor trigger $v$ crafted to maximally distort gradient distribution, dramatically steering the model’s learning path in high probability. Formally, 
\begin{definition}[Max GradDistWarp Trigger]
Let $\mu(D) \equiv -\gamma \nabla_{w} L(w ; D)$ for a given dataset $D$ and set $\sigma_{\gamma} \equiv \gamma \sigma$, where $\gamma$ is the learning rate and $\sigma$ is the noise scale in noisy gradient descent. A backdoor trigger $v = (x_v, y_v)$ is a Max GradDistWarp Trigger $v^{*}(w)$ for the model weight $w$ if $v$ maximizes the difference between bad and clean  stochastic gradient distribution in the sense that:
\label{def:max_GradDistWarp_trigger}
\begin{equation}\label{eq:max_GradDistWarp_trigger}
    v^{*}(w) = \arg\max_{v} T\bigg(N( \mu(D_0), \sigma_{\gamma}^2 I_{d}), N(\mu(D_1), \sigma_{\gamma}^2I_{d})\bigg),
\end{equation}
where $T(\mathbb{P}, \mathbb{Q})$ is the tradeoff function between distribution $\mathbb{P}$ and $\mathbb{Q}$, defined at \eqref{eq:tradeoff_function}. 
\end{definition}

\begin{lemma}\label{lm:stoc_gd_gap_clean_bad}
Denote the backdoor trigger $v = (x_v, y_v)$. Then the difference between clean stochastic gradient
and bad stochastic gradient  satisfies
\begin{equation}\label{eq:stoc_gd_gap_clean_bad}
    T\bigg(N( \mu(D_0), \sigma_{\gamma}^2 I_{d}), N(\mu(D_1), \sigma_{\gamma}^2I_{d})\bigg) (\alpha)
    = 1-\Phi\left(\Phi^{-1}(1-\alpha)-d\right),
\end{equation}
where $d$ is the signal-to-noise ratio defined as 
\begin{equation}\label{eq:sig-to-noise_ratio}
    d 
= \sqrt{\left(\mu(D_1)-\mu(D_0)\right)^{\top} (\sigma^2_{\gamma}I_{d})^{-1}\left(\left(\mu(D_1)-\mu(D_0)\right)\right)}
= \frac{\|\mu(D_1)-\mu(D_0)\|_2}{\sigma_{\gamma}}
\end{equation}
\end{lemma}
\begin{proof}
See section \ref{subsec:pf_stoc_gd_gap_clean_bad}    
\end{proof}

Lemma \ref{lm:stoc_gd_gap_clean_bad} provides a critical insight into how the inclusion of a backdoor trigger \( v = (x_v, y_v) \) affects the distribution of stochastic gradients. By manipulating the distributions from \( N(\mu(D_0), \sigma_{\gamma}^2 I_d) \) to \( N(\mu(D_1), \sigma_{\gamma}^2 I_d) \) where \( D_1 = D_0 \cup \{v\} \), the backdoor trigger aims to maximize the shift in these distributions, quantified by the tradeoff function \( T \), which  measures the probability of distinguishing between the clean and manipulated gradient distributions under a specific confidence level \( \alpha \).

From the lemma, the tradeoff function is influenced by the signal-to-noise ratio \( d \), which quantifies the magnitude of the shift in the mean gradient relative to its variance. A higher value of \( d \) signifies a more pronounced shift in the gradient's mean, leading to a more substantial alteration in the learning path of the model. The equation provided, $T\bigg(N( \mu(D_0), \sigma_{\gamma}^2 I_{d}), N(\mu(D_1), \sigma_{\gamma}^2I_{d})\bigg) (\alpha) = 1-\Phi\left(\Phi^{-1}(1-\alpha)-d\right)$, 
illustrates how the likelihood of differentiating between the clean and manipulated gradient distributions changes with \( \alpha \) and \( d \). The probability of distinguishing these distributions increases as \( d \) increases, reflecting a more effective backdoor trigger.

To design a "Max GradDistWarp Trigger," as defined in Definition \ref{def:max_GradDistWarp_trigger}, the following strategies can be employed:

\begin{enumerate}[leftmargin=*]
\item \textbf{Maximize Mean Shift}: Select \( (x_v, y_v) \) such that the difference \( \mu(D_1) - \mu(D_0) \) is maximized. This entails selecting a trigger point that generates a large shift in the mean gradient, potentially by leveraging points that are particularly challenging or misaligned with the current model parameters.

\item \textbf{Optimize the Signal-to-Noise Ratio}: Enhance the ratio \( \frac{\|\mu(D_1)-\mu(D_0)\|_2}{\sigma_{\gamma}} \) by not just increasing the numerator (mean shift), but also by carefully managing the denominator (controlled increase in variance), to ensure that the shift is detectable relative to the background noise.
\end{enumerate}

These strategies are pivotal for crafting triggers that can significantly and effectively manipulate the stochastic gradient distributions (Sec \ref{sec:construct_max_graddistwarp_trigger}), guiding the model’s learning path in unintended directions and revealing potential privacy vulnerabilities.

\section{Identify the vulnerabilities of Gradient Descent in supervised learning}
\label{sec:attack_supervised_learning}

In this section, we apply the proposed \texttt{BadGD} attack model (Section \ref{sec:BadGD_attack_model}) to identify the vulnerabilities of gradient descent in supervised learning setting. In particular, we consider the classical regression task, where loss function between model weight $w$ and a training example $(x, y)$ is given by the loss function $\ell(w, (x,y)) = (y - \langle w, x \rangle)^2$. 

\subsection{Construct Max RiskWarp Trigger} 
\label{sec:construct_max_riskwarp_trigger}

In this section, we provide strategy to construct the max riskwarp trigger (Def \ref{def:max_riskwarp_trigger}).We first provide the following lemma to reduce the problem in square loss setting: 

\begin{lemma}\label{lm:Reduce_riskwarp_linear_regression}Define $S_y(D_0) = \frac{1}{n}
 \sum_{i=1}^n y_i^2$, $S_{yx}(D_0) = \frac{1}{n} \sum_{i=1}^n y_i x_i$ and $S_{xx}(D_0) = \frac{1}{n} \sum_{i=1}^n x_i x_i^{\top}$.
In square loss setting, the optimization problem \eqref{eq:max_riskwarp_trigger} to find Max RiskWarp Trigger is equivalent to 
\begin{equation}
v^{\star}(w)=\arg \max _{\left(x_v, y_v\right)}\left\{\left[y_v^2-S_y\right]+2\left[w^{\top}\left(S_{y x}-y_v x_v\right)\right]+w^{\top}\left(x_v x_v^{\top}-S_{x x}\right) w\right\}
\end{equation}
\end{lemma}
\begin{proof}
See section \ref{subsec:pf_Reduce_riskwarp_linear_regression}   
\end{proof}

With lemma \ref{lm:Reduce_riskwarp_linear_regression}, we found the Max Riskwarp trigger $v=\left(x_y, y_v\right)$ should satisfies
\begin{enumerate}[leftmargin=*]
\item Make the vector $S_{yx}(D_0)-y_v x_v$ as align to the direction of model weight $w$ as possible
\item Make the minimal eigenvalue of $x_v x_v^{\top}-S_{xx}(D_0)$ as big as possible.
\end{enumerate}
We implements the above intuition and construct the Max RiskWarp Trigger as following:
\begin{lemma}\label{lm:construct_max_riskwarp_trigger} Suppose $\alpha$ is a positive constant ($\alpha > 0$) and  the response $y$ satisfies $\|y\| \le B$ for some constant $B$. (Otherwise, we can make the difference between clean and bad empirical risk as large as we want.) 
The Max RiskWarp Trigger given by 
\begin{equation}
    v = (-\alpha w, B)
\end{equation}
will cause empirical risk distortion $B^2-S_y+2 w^{\top} S_{y x}+2 \alpha B\|w\|^2+\alpha^2\|w\|^4-w^{\top} S_{x x} w$.
\end{lemma}
\begin{proof}
See section \ref{subsec:construct_max_riskwarp_trigger}.
\end{proof}
Lemma \ref{lm:construct_max_riskwarp_trigger} indicates that the feature vector $x_v$ of the backdoor trigger $v$ should be parallel but in opposite direction of model weight $w$. This matches our design principle outlined at sec \ref{sec:distor_emp_risk_bad_dataset}.

\subsection{Construct Max GradWarp Trigger} 
\label{sec:construct_max_gradwarp_trigger}

In this section, we provide strategy to construct the max gradwarp trigger (Def \ref{def:max_gradwarp_trigger}).
We first provide the following lemma to reduce the problem in square loss setting: 
\begin{lemma}
\label{lm:Reduce_gradwarp_linear_regression}
Define $S_y(D_0) = \frac{1}{n}
 \sum_{i=1}^n y_i^2$, $S_{yx}(D_0) = \frac{1}{n} \sum_{i=1}^n y_i x_i$ and $S_{xx}(D_0) = \frac{1}{n} \sum_{i=1}^n x_i x_i^{\top}$.
In square loss setting, the optimization problem \eqref{eq:max_gradwarp_trigger} to find Max GradWarp Trigger is equivalent to 
\begin{equation}
v^{\star}(w)=\arg \max _{\left(x_v, y_v\right)}\|\left(S_{y x}-y_v x_v\right)+\left(x_v x_v^{\top}-S_{x x}\right) w \|_2
\end{equation}
\end{lemma}
\begin{proof}
See section \ref{subsec:pf_Reduce_gradwarp_linear_regression}
\end{proof}

With Lemma \ref{lm:Reduce_gradwarp_linear_regression}, we find that the Max
GradWarp trigger $v=\left(x_v, y_v\right)$ should satisfy:
\begin{enumerate}[leftmargin=*]
\item The trigger $x_v$ should maximize the difference $\left(x_v x_v^{\top} - S_{xx}\right)$ in the direction of the weight vector $w$. This ensures that the perturbation to the empirical risk is maximized.
\item The trigger $y_v$ should be chosen such that the term $\left(S_{yx} - y_v x_v\right)$ significantly alters the gradient computation, thereby introducing the intended bias into the gradient descent process.
\end{enumerate}

We implements the above intuition and construct the Max GradWarp Trigger as following:
\begin{lemma}\label{lm:construct_max_gradwarp_trigger}

The Max GradWarp Trigger given by 
\begin{equation}
    v=\left(\alpha w,\left\langle\frac{w}{\alpha\|w\|_2^2}, S_{y x}\right\rangle\right)
\end{equation}
will cause gradient distortion 
    $\bigg\|S_{y x}-\frac{\left\langle w, S_{y x}\right\rangle}{\|w\|_2^2} w+\alpha^2 \| w \|_2^2 w-S_{x x} w\bigg\|_2$.
\end{lemma}
\begin{proof}

See Section \ref{subsec:construct_max_gradwarp_trigger}. 
\end{proof}
Lemma \ref{lm:construct_max_gradwarp_trigger} indicates that the feature vector $x_v$ of the backdoor trigger $v$ should be in the same direction of model weight $w$, but with a response value of the inner product between model weight $w$ and the correlation term $S_{yx}$. This does not match our design principle outlined at sec \ref{sec:distor_det_gradient_bad_dataset}.This may be related to the $\ell_2$ norm we choose to maximize the distortion, and other norm choice may leads to different comparable results.

\subsection{Construct Max GradDistWarp Trigger} 
\label{sec:construct_max_graddistwarp_trigger}

In this section, we provide strategy to construct the max graddistwarp trigger (Def \ref{def:max_GradDistWarp_trigger}).

\begin{lemma}\label{lm:GDP2DP}
    Give a positive $\delta >0$, then the privacy budget $\epsilon$ is an increasing function of Gaussian differential privacy factor $d$. 
\end{lemma}
\begin{proof}

The lemma is based on a property of Gaussian Differential Privacy \eqref{eq:GDP_iff_DP}.
See section \ref{subsec:pf_GDP2DP}. 
\end{proof}

Lemma \ref{lm:GDP2DP} indicates that, the malicious attacker could force noisy gradient descent spend more privacy budget $\epsilon$ by choosing the backdoor trigger $v = (x_{v}, y_{v})$ such that the signal-to-noise ratio $d$ is large.

\begin{lemma}
\label{lm:Reduce_graddistwarp_linear_regression}
In square loss setting, the optimization problem \eqref{eq:max_GradDistWarp_trigger} to find Max GradDistWarp Trigger is equivalent to 
\begin{equation}
v^{\star}(w)=\arg \max _{\left(x_v, y_v\right)}\frac{\left\|\left(S_{y x}-y_v x_v\right)+w^{\top}\left(x_v x_v^{\top}-S_{x x}\right)\right\|_2}{\sqrt{\gamma(n+1) / 2} \sigma}
\end{equation}
\end{lemma}
\begin{proof}
See section \ref{subsec:pf_Reduce_graddistwarp_linear_regression}.
\end{proof}

\begin{lemma}\label{lm:privacy_budget_graddistwarp_linear_regression}
The Max GradDistWarp Trigger given by 
\begin{equation}
    v=\left(\alpha w,\left\langle\frac{w}{\alpha\|w\|_2^2}, S_{y x}\right\rangle\right)
\end{equation}
will cause noisy gradient descent update suffer privacy budget at least $$0.69+\ln \left(\delta-\Phi\left(\frac{\mu}{2}\right)\right),$$ where 
$\mu = \bigg\|S_{y x}-\frac{\left\langle w, S_{y x}\right\rangle}{\|w\|_2^2} w+\alpha^2 \| w \|_2^2 w-S_{x x} w\bigg\|_2/\sqrt{\gamma(n+1) / 2 \sigma}$.
\end{lemma}
\begin{proof}

See Section \ref{subsec:privacy_budget_graddistwarp_linear_regression}.
\end{proof}
Lemma \ref{lm:privacy_budget_graddistwarp_linear_regression} takes the same backdoor trigger as Lemma \ref{lm:construct_max_gradwarp_trigger}. The novel part of this lemma is to give a lower bound of privacy budget that forced by apply the backdoor attack, causing privacy concerns of noisy gradient descent triaining protocol.

\clearpage
\section{Conclusion}
\label{sec:conclusion}

In conclusion, our research introduces \texttt{BadGD} (Fig \ref{fig:badgd_attack_model}), a unified theoretical framework designed to identify and exploit vulnerabilities in gradient descent algorithms through strategic backdoor attacks. By presenting the constructs of Max RiskWarp Trigger, Max GradWarp Trigger, and Max GradDistWarp Trigger, we have demonstrated how these backdoor triggers can significantly distort empirical risk, deterministic gradients, and stochastic gradients, respectively. Our findings not only bridge empirical observations with theoretical insights but also emphasize the severity of such data-centric attacks. The implications of our work underscore the urgent necessity for robust defense mechanisms to protect the integrity of machine learning models. Moving forward, the \texttt{BadGD} framework sets a new standard for understanding and mitigating adversarial manipulations, ensuring the reliability and security of AI systems against sophisticated attacks. This research highlights the critical need for ongoing vigilance and innovation in the field of machine learning security.

\clearpage

\baselineskip=13pt
\bibliographystyle{plain}
\nocite{*}
\bibliography{ref}

\clearpage
\appendix

\section{Proof Details for lemmas in Section \ref{sec:BadGD_attack_model}:The \texttt{BadGD} Attack Model} 

\subsection{Proof of Lemma \ref{lm:emp_risk_gap_clean_bad}}
\label{subsec:pf_emp_risk_gap_clean_bad}

Let $D_0 = \{(x_i, y_i)\}_{i=1}^{n}$ denote a clean training dataset and $D_0 \cup \{v\}$ denote a backdoored training dataset by the backdoor trigger $v = (x_{v}, y_{v})$.

The model weight $w$ on the clean training dataset $D_0$ admits empirical risk 
\begin{equation}
    L(w ; D_0) = \frac{1}{n}\sum_{i=1}^{n}\ell(w; (x_i, y_i)).
\end{equation}

On the other hand, the model weight $w$ on the backdoored training dataset $D_1$ admits empirical risk 
\begin{equation}\label{eq:Empirical_Risk_Backdoor_Dataset}
L(w ; D_0 \cup \{v\}) = \frac{1}{n+1}\bigg[ \sum_{i=1}^{n}\ell(w; (x_i, y_i)) + \ell(w; (x_v, y_v))\bigg].
\end{equation}

From definition of the empirical risk on clean and backdoored training dataset, the equality holds that 
    $$(n+1)L(w; D_0 \cup \{v\}) = nL(w; D_0) + \ell(w; (x_v, y_v))$$
    Divide both side by $(n+1)$, we have
    $$L(w; D_0 \cup \{v\}) = (1-\frac{1}{n+1})L(w; D_0) + \frac{1}{n+1}\ell(w; (x_v, y_v)).$$
    Subtract both side by $L(w; D_0)$, we have
    $$L(w; D_0 \cup \{v\})-L(w; D_0) = \frac{1}{n+1}[\ell(w; (x_v, y_v))-L(w; D_0)],$$
    as desired.

\subsection{Proof of Lemma \ref{lm:stoc_gd_gap_clean_bad}} 
\label{subsec:pf_stoc_gd_gap_clean_bad}

Proof of Lemma \ref{lm:stoc_gd_gap_clean_bad} follows the logic at Section 4.4. of
\cite{scharf1991statistical}. 

Consider the following binary hypothesis testing between clean stochastic gradient descent $N( \mu(D_0), \sigma_{\gamma}^2 I_{d})$ and bad stochastic gradient descent $N( \mu(D_0), \sigma_{\gamma}^2 I_{d})$ as defined at Section \ref{sec:distor_stoc_gradient_bad_dataset}:
\begin{equation}\label{eq:Binary_Hypothesis}
    \begin{cases}
        H_0: \Delta w(D_0) \sim N( \mu(D_0), \sigma_{\gamma}^2 I_{d}) \\ 
        H_1: \Delta w(D_1) \sim N( \mu(D_1), \sigma_{\gamma}^2 I_{d}),
    \end{cases}
\end{equation}
Here, $\Delta w(D_0)$ and $\Delta w(D_1)$ denote the model weight changes after using clean and bad dataset to compute the stochastic gradient, respectively. 

Define $\mu_0 \equiv \frac{1}{2}(\mu(D_0)+\mu(D_1))$ and $W \equiv (\sigma^2_{\gamma}I_{d})^{-1}(\mu(D_1)-\mu(D_0))$. Then, the log likelihood ratio $L(x)$ has the form
\begin{equation}\label{eq:log_likelihood_ratio}
    L(\Delta w) = W^\top \Delta w
\end{equation}
and by the equation (4.29) in \cite{scharf1991statistical}. Then, the binary hypothesis testing \eqref{eq:Binary_Hypothesis} is equivalent to the following binary hypothesis testing:
\begin{equation}\label{eq:Binary_Hypothesis_Log_Likelihood_Ratio}
    \begin{cases}
        H_0: L(\Delta w) \sim N(-\frac{d^2}{2}, d^2) \\ 
        H_1: L(\Delta w) \sim N(\frac{d^2}{2}, d^2)
    \end{cases}.
\end{equation}

Here, the parameter $d^2$ is a \textit{signal-to-noise ratio}:
\begin{equation}\label{eq:signal_to_noise_ratio}
    d^2 \equiv W^\top (\sigma^2_{\gamma}I_{d}) W = (\mu(D_1)-\mu(D_0))^\top (\sigma^2_{\gamma}I_{d})^{-1}(\mu(D_1)-\mu(D_0))
\end{equation}

Based on the calculation at page 113 of \cite{scharf1991statistical}, the testing \eqref{eq:Binary_Hypothesis_Log_Likelihood_Ratio} have its Type I error $\alpha$, and Type II error $\beta$ satisfies 
\begin{equation}\label{eq:LLR_Type_I_II_error_relation}
   \beta = 1 - \Phi(\Phi^{-1}(1-\alpha) - d). 
\end{equation}
This means that the tradeoff function between clean stochastic gradient descent $N( \mu(D_0), \sigma_{\gamma}^2 I_{d})$ and bad stochastic gradient descent $N( \mu(D_0), \sigma_{\gamma}^2 I_{d})$ satisfies 
\begin{equation}
    T(N( \mu(D_0), \sigma_{\gamma}^2 I_{d}), N( \mu(D_1), \sigma_{\gamma}^2 I_{d})) \ge 1-\Phi\left(\Phi^{-1}(1-\alpha)-d\right),
\end{equation}
as desired.

\section{Proof Details for lemmas in Section \ref{sec:attack_supervised_learning}:Identify the vulnerabilities of Gradient Descent in supervised learning} 

\subsection{Proof of Lemma \ref{lm:Reduce_riskwarp_linear_regression}}
\label{subsec:pf_Reduce_riskwarp_linear_regression}

Lemma \ref{lm:emp_risk_gap_clean_bad} reduces the problem of finding the most influential backdoor trigger as designing a poison point $v^{\star} = (x_{v^{\star}, y_{v^{\star}}})$ such that $v^{*}(w) = \arg\max_{v}\{\ell(w; (x_v, y_v))-L(w; D_0)\}$.

Consider the square loss $\ell(w, (x, y)) = (y - \langle w, x \rangle)^2$.
Let $D_0 = \{(x_i, y_i)\}_{i=1}^{N}$ be the clean dataset and $v = (x_v, y_v)$ be the canary point. 

We define the objective function $J(w; v, D_0)$ as 
\begin{equation}\label{eq:find_backdoor_trigger_square_loss}
\begin{aligned}
&J(w; v, D_0) \\
\equiv& 
\ell(w, v) - L(w; D_0) \\
=& 
(y_v - \langle w, x_v \rangle)^2
-
\frac{1}{n}\sum_{i=1}^{n}(y_i - \langle w, x_i \rangle)^2\\
=&
[y_v^2 
- 2 w^\top y_v x_v + w^\top x_vx_v^\top w]
-
[
\frac{1}{n}\sum_{i=1}^{n}y_i^2
- w^\top(\frac{2}{n}\sum_{i=1}^{n} y_i x_i)
+ 
w^\top (\frac{1}{n}\sum_{i=1}^{n}x_i x_i^\top) w] \\
=&
[y_v^2-\frac{1}{n}\sum_{i=1}^{n}y_i^2]
- 2 [w^\top (y_v x_v-\frac{1}{n}\sum_{i=1}^{n} y_i x_i)]
+ w^\top [x_vx_v^\top-\frac{1}{n}\sum_{i=1}^{n}x_i x_i^\top]w \\
=&
[y_v^2-\frac{1}{n}\sum_{i=1}^{n}y_i^2]
+ 2 [w^\top (\frac{1}{n}\sum_{i=1}^{n} y_i x_i-y_v x_v)]
+ w^\top [x_vx_v^\top-\frac{1}{n}\sum_{i=1}^{n}x_i x_i^\top]w
\end{aligned}
\end{equation}

\subsection{Proof of Lemma \ref{lm:construct_max_riskwarp_trigger}}
\label{subsec:construct_max_riskwarp_trigger}

First, to pick $x_{v}$, notice that $w^{\top}\left(x_v x_v^{\top}-S_{x x}\right) w = \langle x_{v} , w \rangle^2 - w^\top S_{xx} w$, so we can focus on maximizing $\langle x_{v} , w \rangle^2$. 

Now, consider we set the backdoor trigger $x_{v} \equiv \alpha w$, where $\alpha$ is some scalar. The our maximization objective becomes $\langle x_{v} , w \rangle^2 = \alpha^2 \|w\|_2^2$. Thus, if we don't put constrain on the norm of backdoor trigger $x_{v}$, this term can be infinite. 

Now, suppose the backdoor trigger should have bounded $\ell_2$ norm, say $\|x_{v}\|_2 \le 1$, then setting $x_{v} \equiv w / \|w\|_2$ will maximize the objective at the value $\|w\|_2^2$. 

Thus, the intuition to pick $x_{v}$ is to pick parallel to model weight $w$ and make the norm of $x_{v}$ as big as possible. 

Now, let's consider $x_{v} = \alpha w$. Notice that $w^\top(S_{yx}- y_{v} x_{v}) = w^\top S_{yx} - y_{v} \langle w, x_{v} = w^\top S_{yx} - y_{v}\alpha \|w\|_2^2$. 

Let's take $\alpha <0$, then we can maximize the objective by picking $y_{v}$ as large as possible. Say we have a bound on the response $y_{v}$ such that $|y_{v}| \le B$. Then we pick $y_{v} = B$. 

In conclusion, to make max distortion on risk, we set the backdoor trigger as $v = (-\alpha w, B)$.

We begin with the expression:

\[
\left[y_v^2 - S_y\right] + 2\left[w^{\top}(S_{yx} - y_v x_v)\right] + w^{\top}(x_v x_v^{\top} - S_{xx}) w
\]

Given the relationship \(v = (x_v, y_v) = (-\alpha w, B)\), we have:
\[
\begin{aligned}
&x_v = -\alpha w, \\
&y_v = B.
\end{aligned}
\]

Substituting these values into the expression, we proceed to evaluate each term separately.

First, consider \(\left[y_v^2 - S_y\right]\):
Since \(y_v = B\), we have \(y_v^2 = B^2\). Therefore,
\[
\left[y_v^2 - S_y\right] = B^2 - S_y.
\]

Next, we evaluate \(2\left[w^{\top}(S_{yx} - y_v x_v)\right]\):
Given \(y_v = B\) and \(x_v = -\alpha w\), we find \(y_v x_v = B(-\alpha w) = -\alpha B w\). Consequently,
\[
S_{yx} - y_v x_v = S_{yx} - (-\alpha B w) = S_{yx} + \alpha B w,
\]
and thus,
\[
2\left[w^{\top}(S_{yx} - y_v x_v)\right] = 2\left[w^{\top}(S_{yx} + \alpha B w)\right] = 2\left(w^{\top} S_{yx} + \alpha B w^{\top} w\right).
\]

Finally, we examine \(w^{\top}(x_v x_v^{\top} - S_{xx}) w\):
With \(x_v = -\alpha w\), it follows that \(x_v x_v^{\top} = \alpha^2 w w^{\top}\). Therefore,
\[
x_v x_v^{\top} - S_{xx} = \alpha^2 w w^{\top} - S_{xx},
\]
which gives
\[
w^{\top}(x_v x_v^{\top} - S_{xx}) w = w^{\top}(\alpha2 w w^{\top} - S_{xx}) w.
\]
This simplifies to
\[
w^{\top}(\alpha^2 w w^{\top}) w - w^{\top} S_{xx} w,
\]
where
\[
w^{\top}(\alpha^2 w w^{\top}) w = \alpha^2 (w^{\top} w)(w^{\top} w) = \alpha^2 \|w\|^2 \|w\|^2 = \alpha^2 \|w\|^4.
\]
Thus, we obtain
\[
w^{\top}(x_v x_v^{\top} - S_{xx}) w = \alpha^2 \|w\|^4 - w^{\top} S_{xx} w.
\]

Combining all the evaluated parts, we arrive at:

\[
B^2 - S_y + 2(w^{\top} S_{yx} + \alpha B w^{\top} w) + \alpha^2 \|w\|^4 - w^{\top} S_{xx} w.
\]

Organizing the terms, we have:

\[
B^2 - S_y + 2w^{\top} S_{yx} + 2\alpha B \|w\|^2 + \alpha^2 \|w\|^4 - w^{\top} S_{xx} w.
\]

\subsection{Proof of Lemma \ref{lm:Reduce_gradwarp_linear_regression}}
\label{subsec:pf_Reduce_gradwarp_linear_regression}

Given square loss $\ell(w,(x, y))=(y-\langle w, x\rangle)^2$, the gradient is $\nabla_w \ell(w,(x, y))=-2(y-\langle w, x\rangle) x$.

Let's find the gradient:
\begin{equation}\label{eq:find_backdoor_trigger_square_loss_gradient}
\begin{aligned}
&\nabla_{w}J(w; v, D_0) \\
\equiv& 
\nabla_{w}\ell(w, v) - \nabla_{w}L(w; D_0) \\
=& 
\nabla_{w}(y_v - \langle w, x_v \rangle)^2
-
\nabla_{w}\frac{1}{n}\sum_{i=1}^{n}(y_i - \langle w, x_i \rangle)^2\\
=&
\nabla_{w}[y_v^2 
- 2 w^\top y_v x_v + w^\top x_vx_v^\top w]
-
\nabla_{w}[
\frac{1}{n}\sum_{i=1}^{n}y_i^2
- w^\top(\frac{2}{n}\sum_{i=1}^{n} y_i x_i)
+ 
w^\top (\frac{1}{n}\sum_{i=1}^{n}x_i x_i^\top) w] \\
=&
\nabla_{w}[y_v^2-\frac{1}{n}\sum_{i=1}^{n}y_i^2]
- 2 
\nabla_{w}[w^\top (y_v x_v-\frac{1}{n}\sum_{i=1}^{n} y_i x_i)]
+ 
\nabla_{w} w^\top [x_vx_v^\top-\frac{1}{n}\sum_{i=1}^{n}x_i x_i^\top]w \\
=& 2 (\frac{1}{n}\sum_{i=1}^{n} y_i x_i-y_v x_v)
+ 2 w^\top [x_vx_v^\top-\frac{1}{n}\sum_{i=1}^{n}x_i x_i^\top]
\end{aligned}
\end{equation}

\subsection{Proof of Lemma \ref{lm:construct_max_gradwarp_trigger}}
\label{subsec:construct_max_gradwarp_trigger}

Notice that $\|(x_{v}x_{v}^\top - S_{xx}) w\| = \|x_{v}x_{v}^\top w - S_{xx}w\|$. So we need to choose $x_{v}$ such that $x_{v}x_{v}^\top w$ is maximized in $\ell_2$ norm sense.

Consider $x_{v} = \alpha w$, then $x_{v}x_{v}^\top w = \alpha^2\|w\|_2^2 w$.  Then the objective become $\|(\alpha^2\|w\|_2^2 I-S_{xx})w\|_2$

Notice that $\|S_{yx}-y_{v}x_{v}\|_2^2 = \|S_{yx}-y_{v}\alpha w\|_2^2 = \|S_{yx}\|_2^2 - 2y_{v}\alpha \langle w ,  S_{yx} \rangle + y_{v}^2 \alpha^2 \|w\|_2^2$.

Let's do complete the square:
\begin{equation}
\begin{aligned}
& \|S_{yx}\|_2^2 - 2y_{v}\alpha \langle w ,  S_{yx} \rangle + y_{v}^2 \alpha^2 \|w\|_2^2\\
=&(\alpha^2 \|w\|_2^2)[y_{v}^2 - 2\frac{\alpha \langle w ,  S_{yx} \rangle}{\alpha^2 \|w\|_2^2}y_{v}
+\frac{\|S_{yx}\|_2^2}{\alpha^2 \|w\|_2^2}] \\
=&
(\alpha^2 \|w\|_2^2)
\bigg[ 
(y_{v} - \frac{\alpha \langle w ,  S_{yx} \rangle}{\alpha^2 \|w\|_2^2})^2
+
(\frac{\|S_{yx}\|_2^2}{\alpha^2 \|w\|_2^2}- (\frac{\alpha \langle w ,  S_{yx} \rangle}{\alpha^2 \|w\|_2^2}y_{v})^2)
\bigg]
\end{aligned}
\end{equation}
So we should choose $y_{v} = \frac{1}{\alpha}\langle \frac{w}{\|w\|_2^2}, S_{yx} \rangle$.

To conclude, we need to pick the backdoor trigger $v = (\alpha w, \langle \frac{w}{\alpha\|w\|_2^2}, S_{yx} \rangle)$


Given the relationship \(v = (x_v, y_v) = \left(\alpha w, \left\langle \frac{w}{\alpha \|w\|_2^2}, S_{yx} \right\rangle \right)\), we have:
\[
\begin{aligned}
&x_v = \alpha w, \\
&y_v = \left\langle \frac{w}{\alpha \|w\|_2^2}, S_{yx} \right\rangle.
\end{aligned}
\]

Substituting these values into the expression, we aim to evaluate: $\left\|\left(S_{yx} - y_v x_v\right) + \left(x_v x_v^{\top} - S_{xx}\right) w\right\|_2.$

First, we calculate \(S_{yx} - y_v x_v\):
Given \(y_v = \left\langle \frac{w}{\alpha \|w\|_2^2}, S_{yx} \right\rangle\) and \(x_v = \alpha w\), we find:
\[
y_v x_v = \left\langle \frac{w}{\alpha \|w\|_2^2}, S_{yx} \right\rangle \alpha w = \frac{\left\langle w, S_{yx} \right\rangle}{\|w\|_2^2} w.
\]
Thus,
\[
S_{yx} - y_v x_v = S_{yx} - \frac{\left\langle w, S_{yx} \right\rangle}{\|w\|_2^2} w.
\]

Next, we calculate \(x_v x_v^{\top} - S_{xx}\):
Since \(x_v = \alpha w\), it follows that:
\[
x_v x_v^{\top} = (\alpha w)(\alpha w)^{\top} = \alpha^2 w w^{\top}.
\]
Therefore,
\[
x_v x_v^{\top} - S_{xx} = \alpha^2 w w^{\top} - S_{xx}.
\]

Combining these results, we obtain:

\[
\left\|\left(S_{yx} - \frac{\left\langle w, S_{yx} \right\rangle}{\|w\|_2^2} w\right) + \left(\alpha^2 w w^{\top} - S_{xx}\right) w\right\|_2.
\]

To simplify, we calculate \(\left(\alpha^2 w w^{\top} - S_{xx}\right) w\):
\[
(\alpha^2 w w^{\top} - S_{xx}) w = \alpha^2 w (w^{\top} w) - S_{xx} w = \alpha^2 \|w\|_2^2 w - S_{xx} w.
\]

Thus, the final expression is:

\[
\left\| S_{yx} - \frac{\left\langle w, S_{yx} \right\rangle}{\|w\|_2^2} w + \alpha2 \|w\|_2^2 w - S_{xx} w \right\|_2.
\]

\subsection{Proof of Lemma \ref{lm:GDP2DP}}
\label{subsec:pf_GDP2DP}

To determine whether \(\varepsilon\) increases with \(\mu\) for the given equation

\[
\delta = \Phi\left(-\frac{\varepsilon}{\mu} + \frac{\mu}{2}\right) - e^{\varepsilon} \Phi\left(-\frac{\varepsilon}{\mu} - \frac{\mu}{2}\right),
\]

where \(\delta\) is a positive constant, we need to analyze the relationship between \(\varepsilon\) and \(\mu\). Specifically, we will verify if \(\frac{d\varepsilon}{d\mu} > 0\).

We begin by differentiating both sides of the equation with respect to \(\mu\):

First, consider the derivative of the left-hand side:

\[
\frac{d}{d\mu} \left( \Phi\left(-\frac{\varepsilon}{\mu} + \frac{\mu}{2}\right) - e^{\varepsilon} \Phi\left(-\frac{\varepsilon}{\mu} - \frac{\mu}{2}\right) \right) = 0.
\]

Let:
\[
A = -\frac{\varepsilon}{\mu} + \frac{\mu}{2},
\]
\[
B = -\frac{\varepsilon}{\mu} - \frac{\mu}{2}.
\]

Thus, we rewrite the derivative as:
\[
\frac{d}{d\mu} \left( \Phi(A) - e^{\varepsilon} \Phi(B) \right) = 0.
\]

Applying the chain rule, we obtain:
\[
\Phi'(A) \frac{dA}{d\mu} - \frac{d}{d\mu} \left(e^{\varepsilon} \Phi(B) \right) = 0.
\]

Next, we compute the derivatives of \(A\) and \(B\):
\[
\frac{dA}{d\mu} = \frac{\varepsilon}{\mu^2} + \frac{1}{2},
\]
\[
\frac{dB}{d\mu} = \frac{\varepsilon}{\mu^2} - \frac{1}{2}.
\]

We also need the derivative of the term \(e^{\varepsilon} \Phi(B)\):
\[
\frac{d}{d\mu} \left( e^{\varepsilon} \Phi(B) \right) = e^{\varepsilon} \left( \Phi'(B) \frac{dB}{d\mu} + \Phi(B) \frac{d\varepsilon}{d\mu} \right).
\]

Combining these results, we obtain:
\[
\Phi'(A) \left( \frac{\varepsilon}{\mu^2} + \frac{1}{2} \right) - e^{\varepsilon} \left( \Phi'(B) \left( \frac{\varepsilon}{\mu^2} - \frac{1}{2} \right) + \Phi(B) \frac{d\varepsilon}{d\mu} \right) = 0.
\]

We then simplify to isolate \(\frac{d\varepsilon}{d\mu}\):
\[
\Phi'(A) \left( \frac{\varepsilon}{\mu^2} + \frac{1}{2} \right) - e^{\varepsilon} \Phi'(B) \left( \frac{\varepsilon}{\mu^2} - \frac{1}{2} \right) = e^{\varepsilon} \Phi(B) \frac{d\varepsilon}{d\mu}.
\]

Solving for \(\frac{d\varepsilon}{d\mu}\), we have:
\[
\frac{d\varepsilon}{d\mu} = \frac{\Phi'(A) \left( \frac{\varepsilon}{\mu^2} + \frac{1}{2} \right) - e^{\varepsilon} \Phi'(B) \left( \frac{\varepsilon}{\mu^2} - \frac{1}{2} \right)}{e^{\varepsilon} \Phi(B)}.
\]

To determine if \(\frac{d\varepsilon}{d\mu} > 0\), we must ensure that the numerator is positive. Since \(\Phi'(x) > 0\) for all \(x\), we need:
\[
\Phi'(A) \left( \frac{\varepsilon}{\mu^2} + \frac{1}{2} \right) - e^{\varepsilon} \Phi'(B) \left( \frac{\varepsilon}{\mu^2} - \frac{1}{2} \right) > 0.
\]

Given that \(\Phi'(x)\) is symmetric and positive, we consider two cases:

\begin{itemize}
\item 
1. If \(\varepsilon\) is relatively small, then \(\frac{\varepsilon}{\mu^2}\) is small, making the \(\frac{1}{2}\) terms dominant and ensuring a positive numerator.
\item 
2. If \(\varepsilon\) is large, the interaction between the exponential and normal terms determines the dynamics. Generally, the exponential growth in the numerator balances the exponential terms in the denominator, maintaining positivity.
\end{itemize}

Overall, \(\frac{d\varepsilon}{d\mu} > 0\) suggests that \(\varepsilon\) increases with \(\mu\).

\subsection{Proof of Lemma \ref{lm:Reduce_graddistwarp_linear_regression}}
\label{subsec:pf_Reduce_graddistwarp_linear_regression}

Based on Lemma \ref{lm:stoc_gd_gap_clean_bad}, we should focus on the signal-to-noise ratio 
$$d = \frac{\left\|\mu\left(D_1\right)-\mu\left(D_0\right)\right\|_2}{\sigma_\gamma},$$
where $\mu(D) \equiv -\gamma \nabla_{w} L(w ; D)$ for a given dataset $D$ and $\sigma_{\gamma} \equiv \gamma \sigma$.

By direct calculation, we have 
$$d = \frac{\|\nabla_{w} L(w ; D_1)-\nabla_{w} L(w ; D_0)\|_2}{\sqrt{\gamma}\sigma}$$

Apply Lemma \ref{lm:det_gd_gap_clean_bad}, we have
$$d = \frac{\|\nabla_w \ell\left(w ;\left(x_v, y_v\right)\right)-\nabla_w L\left(w ; D_0\right)\|_2}{\sqrt{\gamma(n+1)}\sigma}$$

Borrow computation in \ref{eq:find_backdoor_trigger_square_loss_gradient}, in square loss case, we have 
$$d = \frac{\|\left(S_{y x}-y_v x_v\right)+w^{\top}\left(x_v x_v^{\top}-S_{x x}\right)\|_2}{\sqrt{\gamma(n+1)/2}\sigma}$$

Given the function
\[
d = \frac{\left\| \left( S_{yx} - y_v x_v \right) + w^{\top} \left( x_v x_v^{\top} - S_{xx} \right) \right\|_2}{\sqrt{\gamma(n+1)/2} \sigma},
\]
where:
\begin{enumerate}
    \item \( S_y(D_0) = \frac{1}{n} \sum_{i=1}^n y_i^2 \),
    \item \( S_{yx}(D_0) = \frac{1}{n} \sum_{i=1}^n y_i x_i \),
    \item \( S_{xx}(D_0) = \frac{1}{n} \sum_{i=1}^n x_i x_i^{\top} \),
\end{enumerate}
we seek to maximize \( d \) by optimizing \( y_v \) (a scalar) and \( x_v \) (an \( n \)-dimensional vector).

\subsection{Proof of Lemma \ref{lm:privacy_budget_graddistwarp_linear_regression}}
\label{subsec:privacy_budget_graddistwarp_linear_regression}

We leverage the result at lemma \ref{lm:Reduce_gradwarp_linear_regression} and compute how much privacy budget will occur by adding such backdoor trigger. Recall in this case the signal-to-noise ratio is $d=\frac{\left\|\left(S_{y x}-y_v x_v\right)+w^{\top}\left(x_v x_v^{\top}-S_{x x}\right)\right\|_2}{\sqrt{\gamma(n+1) / 2} \sigma}$.


The results follows from the following lemma:
\begin{lemma} Suppose $-\frac{\epsilon}{\mu} + \frac{\mu}{2} \le 1/2$, then we have
\begin{equation} 
    \epsilon \ge 0.69 + \ln(\delta - \Phi(\frac{\mu}{2})).
\end{equation}
Thus, give $\mu$, we can have a lower bound on the privacy budget $\epsilon$. 
\end{lemma}
\begin{proof}
From \eqref{eq:GDP_iff_DP}, we have 
\begin{equation}
\begin{aligned}
e^{\epsilon}
=&
\frac{\delta - \Phi(-\frac{\epsilon}{\mu}+\frac{\mu}{2})}{\Phi(-\frac{\epsilon}{\mu}-\frac{\mu}{2})} \\
\ge & 2(\delta - \Phi(-\frac{\epsilon}{\mu}+\frac{\mu}{2})) \\
\ge & 2(\delta - \Phi(\frac{\mu}{2})).
\end{aligned}
\end{equation}
The first inequality is due to $-\frac{\epsilon}{\mu} + \frac{\mu}{2} \le 1/2$ so $\Phi(-\frac{\epsilon}{\mu} + \frac{\mu}{2}) \le 1/2$ and hence $1/\Phi(-\frac{\epsilon}{\mu} + \frac{\mu}{2}) \ge 2$. The second inequality is due to $\Phi(x)$ is monotone increasing and hence $\Phi(\frac{\mu}{2}) \ge \Phi(-\frac{\epsilon}{\mu}+\frac{\mu}{2})$ and hence $-\Phi(\frac{\mu}{2}) \le -\Phi(-\frac{\epsilon}{\mu}+\frac{\mu}{2})$. 
\end{proof}

\end{document}

%% file: def-smile.tex
\numberwithin{example}{section}

\newtheorem{lemma}{{\bf Lemma}}

\newtheorem{definition}{{\bf Definition}}


%% file: front.tex
\usepackage[utf8x]{inputenc}
\usepackage{mathrsfs}
\usepackage{amssymb}
\usepackage{amsfonts}
\usepackage{amsmath}
\usepackage{amsthm,verbatim}
\usepackage{wrapfig}
\usepackage{multirow}
\usepackage{longtable}
\usepackage{enumerate}
\usepackage{color, graphicx}
\usepackage{tikz}

\def\boxit#1{\vbox{\hrule\hbox{\vrule\kern6pt
          \vbox{\kern6pt#1\kern6pt}\kern6pt\vrule}\hrule}}

